\documentclass{iopjournal}

\begin{document}

\articletype{Paper} 

\title{A Surrogate Model for High Temperature Superconducting Magnets to Predict Current Distribution with Neural Networks}

\author{Mianjun Xiao$^1$\orcid{0000-0000-0000-0000}, 
Peng Song$^{1,2}$\orcid{0000-0000-0000-0000}, 
Yulong Liu$^1$\orcid{0000-0000-0000-0000}, 
Cedric Korte$^1$\orcid{0000-0000-0000-0000}, 
Ziyang Xu$^1$\orcid{0000-0000-0000-0000},
Jiale Gao$^1$\orcid{0000-0000-0000-0000}, 
Jiaqi Lu$^1$\orcid{0000-0000-0000-0000}, 
Haoyang Nie$^1$\orcid{0000-0000-0000-0000},
Qiantong Deng$^1$\orcid{0000-0000-0000-0000},
and Timing Qu$^{1,2*}$\orcid{0000-0000-0000-0000}}

\affil{$^1$State Key Laboratory of Clean and Efficient Turbomachinery Power Equipment, Department of Mechanical Engineering, Tsinghua University, Beijing 100084, China}

\affil{$^2$Key Laboratory for Advanced Materials Processing Technology, Ministry of Education, Beijing 100084, China}

\affil{$^*$Author to whom any correspondence should be addressed.}

\email{xmj21@mails.tsinghua.edu.cn}


\begin{abstract}
Finite element methods (FEM) for high-temperature superconducting (HTS) magnets become time-consuming at larger scales, restricting the rapid optimization of meter-scale REBCO solenoids. In this work, a surrogate model based on a fully connected residual neural network (FCRN) is developed to predict the current density distribution in REBCO solenoids. Trained on datasets generated from FEM simulations by the $T-A$ formulation, the FCRN model is evaluated under both fast ramping and steady-state scenarios, showing a lower validation loss than the fully connected network (FCN). When extrapolating geometric parameters beyond the training set, the model achieves a relative error of below 10\% for magnetization losses in Case 1 and an average error of 1.2\% for the central magnetic field in Case 2. Furthermore, deploying the steady-state surrogate model for rapid magnet design found the optimal solution within the parameter space under constraints, with a relative central magnetic field error of 0.2\% compared to FEM results. With rapid predictions, this surrogate model offers an efficient tool for the intelligent design of large-scale HTS magnets.
\end{abstract}

\section{Introduction}
Rare-earth barium copper oxide (REBCO) coated conductors enable field strengths beyond the limits of low-temperature superconductors with higher critical current and upper critical magnetic field \cite{Uglietti2019}. Their applications in large-scale magnet systems include central solenoids in fusion Tokamak \cite{Charlie2024, Hartwig2023}, ultra-high field solenoids \cite{Liu2020, Zhang2024} and superconducting motor \cite{Anne2019, Wu2021}, etc. Under transport current or external magnetic field, REBCO coated conductors exhibit screening current effects, resulting in a nonuniform current distribution inside the tapes \cite{Norris1970}. These effects reduce the central magnetic field compared with the uniform current assumption \cite{Li2019} and induce stress concentrations at the tape ends \cite{Yan2021}, which influence the performance of the magnet. Therefore, the accurate calculation of current density distributions is important for the reliable design and optimization of future high-field magnets.

At present, FEM is widely used for calculating magnetization currents in superconductors. Depending on state variables, these methods include $H$-formulation \cite{Hong2006}, $T-A$ formulation \cite{Zhang2016}, $H-\Phi$ formulation \cite{Arsenault2021}, etc. The computation time of FEM increases with the size of the magnet. For meter-scale HTS magnets, a single operating condition may require tens of hours for FEM computation. In particular, when additional couplings between the electromagnetic field and other physical fields are considered, such as electromagnetic–mechanical or electromagnetic–thermal couplings, the computation may require several hundred hours or even longer \cite{Shao2023, Liu2025}. This computational constraint has become one of the key challenges limiting the fast optimization of high-field REBCO magnet designs.

In recent years, artificial intelligence and machine learning methods have gained increasing attention in applied superconductivity for their rapid prediction abilities. Many studies have demonstrated the feasibility of neural networks for specific tasks, including estimating levitation forces \cite{Liu2022, Ke2022}, computing magnetic fields \cite{Wang2023, Mohammad2023} and AC losses \cite{Zhang2018, Zhou2025} in HTS magnets, predicting material properties \cite{Mohammad2022, Shahriar2025}, and evaluating the critical current ($J_c$) under different field and temperature conditions \cite{Giacomo2022, Zhu2022, Nitish2023}. These studies have demonstrated excellent performance on their specific prediction tasks.

In this work, the focus is placed on HTS solenoids considering the screening current effect, and the model is trained to directly predict the current density distribution of the solenoid, as it directly determines other properties such as magnetic fields, magnetization losses, and electromagnetic force, etc. Such a model can be applied to predict a broader range of HTS magnet characteristics, which is of significant importance.

However, developing a neural network with $J_c$ as outputs meets several challenges. First, the current distribution in REBCO magnets is governed by structure geometries and operation dynamics. Constructing datasets that fully cover these variations leads to massive data sizes. Second, training neural networks on such datasets requires corresponding architecture design to avoid over-fitting or under-fitting. Therefore, selecting suitable network architectures and hyperparameters is essential to balance prediction accuracy, generalization capability, and computational cost.

This study develops a surrogate model based on a fully connected residual network (FCRN) to predict the current density distribution in REBCO solenoids. The model is trained on datasets generated by FEM simulations using the $T$–$A$ formulation. Interpolation accuracy, extrapolation capability, and computational efficiency are systematically evaluated under both fast ramping and steady-state conditions. Results demonstrate that the FCRN model achieves accurate extrapolation and enables rapid magnet design while significantly reducing computation time.

The paper is organized as follows: Section 2 introduces the surrogate model framework, Section 3 describes the FEM dataset generation, Section 4 presents the network architecture and training strategy, Section 5 discusses the predictive performances and a rapid magnet design application based on the surrogate model, and Section 6 provides conclusions and perspectives for future work.

\section{Description of the surrogate model}
\begin{figure}[htbp]
	\centering
	\includegraphics[width=0.95\textwidth]{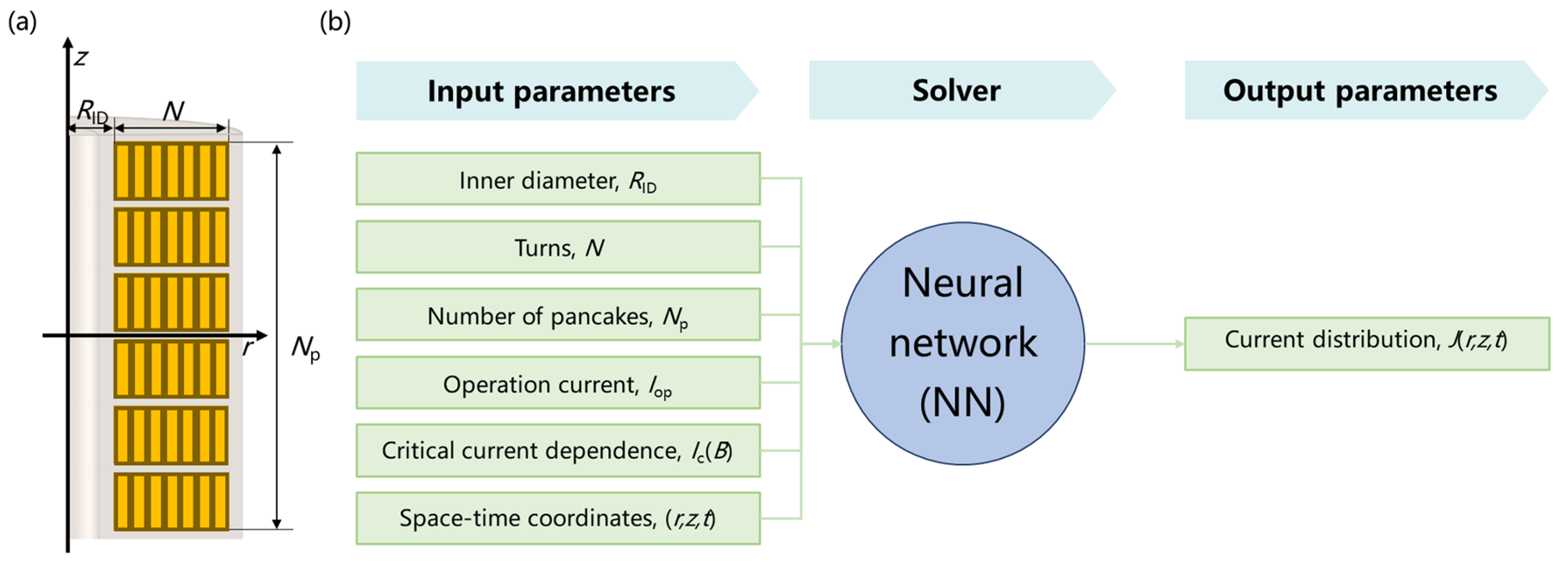}
	\caption{(a)  Geometry of the HTS solenoid, where $R_{\mathrm{ID}}$ denotes the inner diameter, $N$ the number of turns, and $N_{\mathrm{p}}$ the number of pancakes. (b) Schematic of the surrogate model, with inputs representing the HTS solenoid coil parameters (e.g., inner diameter, number of turns and pancakes, operation current) and outputs given as the spatial distribution of the current density.}
	\label{surrogate_model}
\end{figure}

The geometry of the HTS solenoid considered in this study is shown in Fig. \ref{surrogate_model}(a), with a REBCO tape width of 4\:mm and thickness of 0.1\:mm. The schematic of the surrogate model is shown in Fig. \ref{surrogate_model}(b). For an HTS solenoid magnet, the geometric structure is characterized by parameters such as the inner diameter, number of turns and pancakes. The electromagnetic properties are defined by the operating current, ramping rate, and the critical current of HTS tapes with their field dependence. These parameters constitute the inputs of the surrogate model. The solver processes these inputs and computes the solution with numerical or other approaches. To achieve a rapid solution, the solver in this study employs a data-driven neural network that predicts the spatial distribution of the current density inside the magnet based on the given parameters. 

\begin{figure}[htbp]
	\centering
	\includegraphics[width=0.85\textwidth]{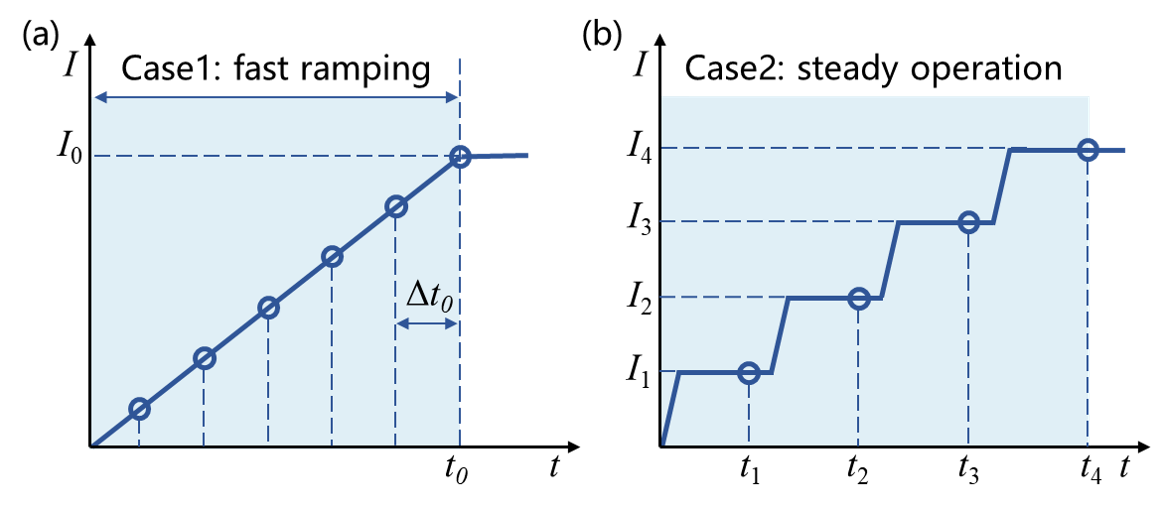}
	\caption{Operating current profiles for the two scenarios considered in this study: (a) fast ramping, and (b) steady operation. Circular markers indicate the time points sampled for  neural network training.}
	\label{case_dif}
\end{figure}

Based on this surrogate model, two distinct operating scenarios are investigated. The first one evaluates the current density distribution of the solenoid during a fast ramping process, as shown in Fig. \ref{case_dif}(a). Because this case requires capturing the dynamic evolution of the magnet, the parameter space is simplified by varying only the number of turns and pancakes, and the magnetic field dependence of the critical current is also neglected. For this fast ramping scenario, the inner diameter is fixed at 10\:mm, the target operating current $I_0$ is 50\:A, the total ramping time $t_0$ is 1\:s, and the time step $\Delta t_0$ is 0.1\:s. Consequently, the neural network learns from data at 10 discrete time points for each individual ramping operation.

The second one represents a steady-state operation condition, to evaluate the current density distribution when the magnet stabilized at its target operating current. Under this situation, the inner diameter, number of turns and pancakes, and the operating current are treated as variables. Additionally, the $J_{\mathrm{c}}(B)$ relationship is considered in this case. For solenoids with different structural parameters, the model predicts the steady-state current density distributions across various operating currents with a stepwise ramping profile shown in Fig. \ref{case_dif}(b).

\section{Datasets generation}
\subsection{Finite element model}
\begin{figure}[htbp]
	\centering
	\includegraphics[width=0.7\textwidth]{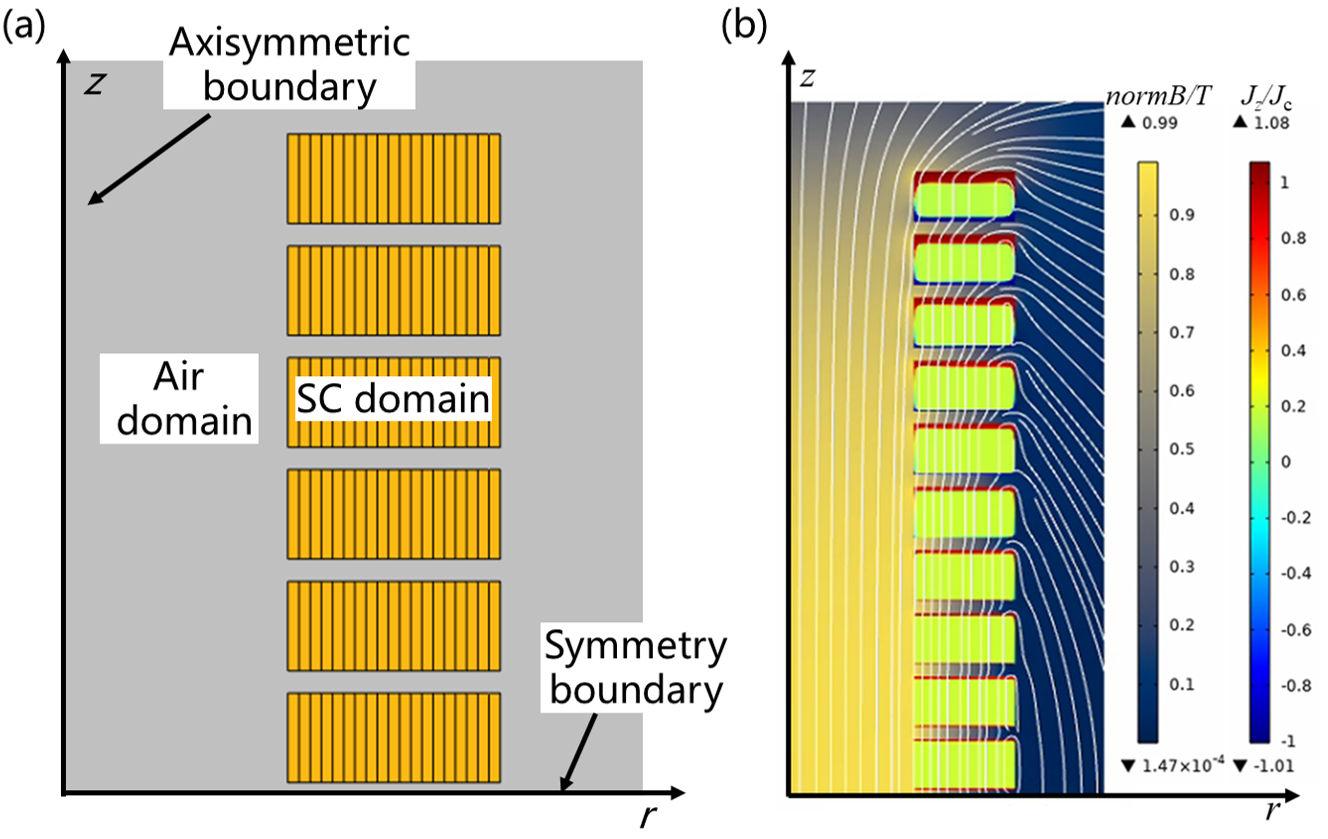}
	\caption{(a) The one-quarter FEM used to reduce computational cost. (b) The current density and magnetic field distributions calculated using the $T$–$A$ formulation.}
	\label{FEM_model}
\end{figure}

The training dataset is generated using a FEM model. To reduce computational cost, a one-quarter model is employed, as shown in Fig. \ref{FEM_model}(a). Symmetry and axisymmetric boundary conditions are applied along the $r$-axis and $z$-axis, respectively. The pancakes in the quarter model are indexed sequentially from bottom to top (1 to $N_{\mathrm{p}}$). Therefore, only solenoids with an even number of total pancakes are considered during training, with the input parameter $N_{\mathrm{p}}$ representing half of the actual total number of pancakes.

The current distribution within the HTS solenoid is calculated by the $T$–$A$ formulation \cite{Zhang2016}, as shown in Fig. \ref{FEM_model}(b). With thin-sheet approximation, the HTS tapes are simplified into two-dimensional lines to reduce computation time. In the superconducting domain, the current vector potential $\mathbf{T}$ is governed by (\ref{eq:T}):
\begin{equation}
	\nabla \times \bigl(\rho\, \nabla \times \mathbf{T}\bigr)= -\,\frac{\partial \mathbf{B}}{\partial t},
	\label{eq:T}
\end{equation}
where $\mathbf{B}$ is the magnetic flux density. In the air domain, the magnetic vector potential $\mathbf{A}$ is chosen as the state variable to solve Ampère’s law (\ref{eq:A}):
\begin{equation}
	\nabla \times \nabla \times \mathbf{A} = \mu_0 \mathbf{J}.
	\label{eq:A}
\end{equation}

The nonlinear electromagnetic behavior of the HTS tapes is characterized by the $E$–$J$ power law (\ref{eq:EJ}):
\begin{equation}
	E = E_{\mathrm{c}} \left(\frac{J}{J_{\mathrm{c}}}\right)^{n-1},
	\label{eq:EJ}
\end{equation}
where the electric-field criterion is $E_{\mathrm{c}}=1\,\mathrm{\mu V/cm}$ and the $n$-value is 21. The critical current density $J_{\mathrm{c}}$ is defined differently for the two operating scenarios. For Case 1 (fast ramping), the magnetic field dependence is neglected, and a constant critical current density $J_{\mathrm{c}}=5\times10^{10}\,\mathrm{A/m^{2}}$ is applied. For Case 2 (steady operation), the magnetic field dependence of the critical current density is characterized by (\ref{eq:JcB}):
\begin{equation}
	J_{\mathrm{c}}(\mathbf{B}) = \frac{J_{\mathrm{c0}}}{\left(1 + \frac{|\mathbf{B}_{\perp}|}{B_0}\right)^\alpha},
	\label{eq:JcB}
\end{equation}
where $J_{\mathrm{c0}}=3.74\times10^{11}\,\mathrm{A/m^{2}}$, $B_0=1.673\,\mathrm{T}$, and $\alpha=0.769$.

\subsection{Post processing}
The dataset is constructed from combinations of the input parameters for the two operating scenarios. The overall parameter ranges are summarized in Table \ref{parameters_dataset}. For both cases, the dataset is divided into a training set and an interpolation validation set.

The training set provides the samples used for loss calculation and back-propagation to update the network weights. For Case 1, the training set consists of 25 parameter combinations, formed by both 5 discrete values of $N$ and $N_{\mathrm{p}}$. For Case 2, the training set comprises 81 parameter combinations, including 3 discrete values for each of the four variables ($N$, $N_{\mathrm{p}}$, $R_{\mathrm{ID}}$ and $I_{\mathrm{op}}$).

The interpolation validation set is used to assess the generalization performance and to select the best-performing model without participating in the network weight updates. The validation samples are generated using intermediate parameter values within the ranges of the training set domain. Specifically, the validation set contains 9 parameter combinations for Case 1 and 16 parameter combinations for Case 2.

\begin{table}[htbp]
	\caption{Dataset Range of the Two Operating Scenarios}
	\centering
	\begin{tabular}{l c c c c}
		\hline
		& \multicolumn{2}{c}{Case 1 (Fast ramping)} & \multicolumn{2}{c}{Case 2 (Steady operation)} \\
		Input parameter & Training set & Validation set & Training set & Validation set \\
		\hline
		$N$ & 10, 30, 50, 80, 100 & 20, 60, 90 & 100, 200, 300 & 150, 250 \\ 
		$N_{\mathrm{p}}$ & 1, 3, 5, 8, 10 & 2, 6, 9 & 2, 5, 8 & 3, 6 \\
		$R_{\mathrm{ID}}$ (mm) & -- & -- & 10, 15, 20 & 12, 17 \\
		$I_{\mathrm{op}}$ (A) & -- & -- & 75, 150, 225 & 110, 190 \\
		\hline
	\end{tabular}
	\label{parameters_dataset}
\end{table}

From the FEM model solutions, the current density within the superconducting domain is extracted at a spatial resolution of 0.1\:mm across each tape. For Case 1, this extraction yields 2.916 million labeled samples for the training set and 1.156 million samples for the interpolation validation set. For Case 2, the training and validation sets contain 3.321 million and 0.8856 million samples, respectively.

Before training, all inputs and outputs are normalized to $[0, 1]$, as detailed in Table \ref{parameters_solenoid}. The neural network requires six input variables for Case 1 ($r$, $z$, $t$, $N$, $N_{\mathrm{p}}$, and $p$) and seven input variables for Case 2 ($r$, $z$, $R_{\mathrm{ID}}$, $I_{\mathrm{op}}$, $N$, $N_{\mathrm{p}}$, and $p$). The variable $p$ acts as an explicit index indicating the respective pancake at the given spatial location. In both operating scenarios, individual pancakes exhibit varied current penetration behaviors due to differences in the local magnetic field. Central pancakes, dominated by parallel field components, exhibit shallower current penetration, whereas pancakes at two ends experience larger perpendicular fields and thus deeper penetration. Including $p$ as an input parameter helps the neural network effectively learn these systematic differences and improves overall model performance.

\begin{table}[htbp]
	\caption{Input and Output Parameters of the Neural Network}
	\centering
	\begin{tabular}{l c c c}
		\hline
		Input parameter & Description & Normalization(Case 1) & Normalization(Case 2)\\
		\hline
		$r$ & radial position & $(r-R_{\mathrm{ID}})*100$ & $(r-R_{\mathrm{ID}})*25$\\ 
		$z$ & axial position & \multicolumn{2}{c}{$(z-5*10^{-4})*250$}\\
		$N$ & number of turns & $N/100$ & $N/300$\\
		$N_{\mathrm{p}}$ & number of pancakes & $N_{\mathrm{p}}/10$ & $N_{\mathrm{p}}/8$\\
		$p$ & the pancake at ($r$,$z$) & $p/10$ & $p/8$\\
		$t$ & time & $t$ & ---\\
		$R_{\mathrm{ID}}$ & inner diameter & --- & $R_{\mathrm{ID}}/20$\\
		$I_{\mathrm{op}}$ & operation current & --- & $I_{\mathrm{op}}/225$\\		
		\hline
		Output parameter\\
		\hline
		$J_{\mathrm{\varphi}}(r,z,t)$ & circumferential current density & \multicolumn{2}{c}{$((J_{\mathrm{\varphi}}/J_{\mathrm{c}})+1)/2$}\\
		\hline
	\end{tabular}
	\label{parameters_solenoid}
\end{table}

\section{Neural Network Structure}
\subsection{Full-connected residual network}
\begin{figure}[htbp]
	\centering
	\includegraphics[width=0.6\textwidth]{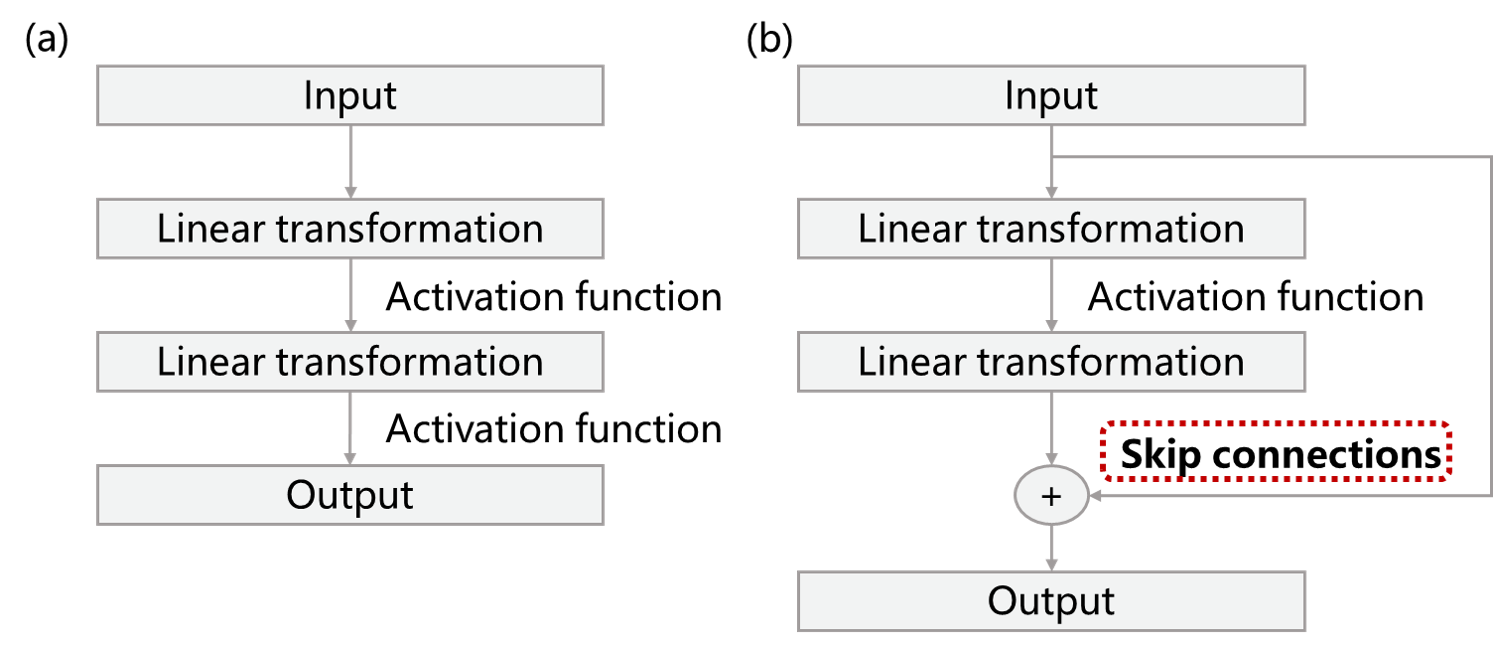}
	\caption{Different neural network structures: (a) conventional fully connected network and (b) residual network with a shortcut connection adding the input directly to the output of the second linear transformation.}
	\label{network_structure}
\end{figure}

The fully connected network is a foundational, widely used and effective network architecture. As shown in figure \ref{network_structure}(a), it maps inputs to outputs by repeatedly applying linear transformations followed by a nonlinear activation function. Due to the large size of the dataset, a neural network with increased depth and width is required. However, traditional FCNs encounter vanishing gradients at greater layers, which can lead to non-convergence of the FCN. To mitigate this, a fully connected residual network was adopted, as shown in figure \ref{network_structure}(b) \cite{He2016}. The key difference is the use of skip connections. Each residual block comprises two linear layers with an activation between them, and the output of the second linear layer is added to the block input without an activation. By introducing skip connections between layers, residual networks create short paths that let signals and gradients propagate directly instead of decaying across many depths, effectively mitigating vanishing gradients.

Table \ref{parameters_network} summarizes the network hyperparameters. The sigmoid linear unit (SiLU) activation function was used as \ref{eq:SiLU}
\begin{equation}
	\mathrm{SiLU}(x)=x \cdot \sigma (x)=\frac{x}{1+e^{-x}},
	\label{eq:SiLU}
\end{equation}
which is smooth with a continuous derivative \cite{Stefan2017}. To compare the effects of different depths and widths, nine kinds of network architectures were trained for both FCN and FCRN. Because each residual block contains two linear layers, the number of residual blocks corresponds to $L/2$ in the FCRN.

For large datasets, training is performed in mini-batches rather than in a single pass. An epoch is defined as one complete pass over all batches, and the training loss for an epoch is computed as the mean of the per-batch losses. In this study, the batch sizes are 4096 for training set and 8192 for validation set. The optimizer of the neural network was the Adam optimizer with an initial learning rate of $5 \times 10^{-4}$, which was reduced to 60\% of its previous value for every 50 epochs. The mean squared error (MSE) loss was adopted for loss evaluations. For a batch containing $N$ samples, the MSE is defined as the average of the squared differences between the predicted values $\hat{y}_i$ and the reference values $y_i$:
\begin{equation}
	\mathrm{MSE} = \frac{1}{N} \sum_{i=1}^{N} \left( \hat{y}_i - y_i \right)^{2}
\end{equation}

In this study, FEM models were calculated on an Intel i9-13900K CPU with 128$\,$GB of memory, and neural network training was carried out on an NVIDIA GeForce RTX 3080 GPU with 12$\,$GB of VRAM.

\begin{table}[htbp]
	\caption{Network hyperparameters}
	\centering
	\begin{tabular}{l c c c}
		\hline
		Input parameter & Description \\
		\hline
		Activation function & SiLU\\ 
		Input dimension & 6(Case1), 7(Case2)\\  
		Output dimension & 1\\
		Number of layers, $L$ & 6,12,24\\
		Neurons per layer, $H$ & 128,256,512\\
		Batch size & 4096(Training set), 8192(Validation set)\\
		Optimizer & Adam\\
		\hline
	\end{tabular}
	\label{parameters_network}
\end{table}

\subsection{Training strategy}
\begin{figure}[htbp]
	\centering
	\includegraphics[width=0.8\textwidth]{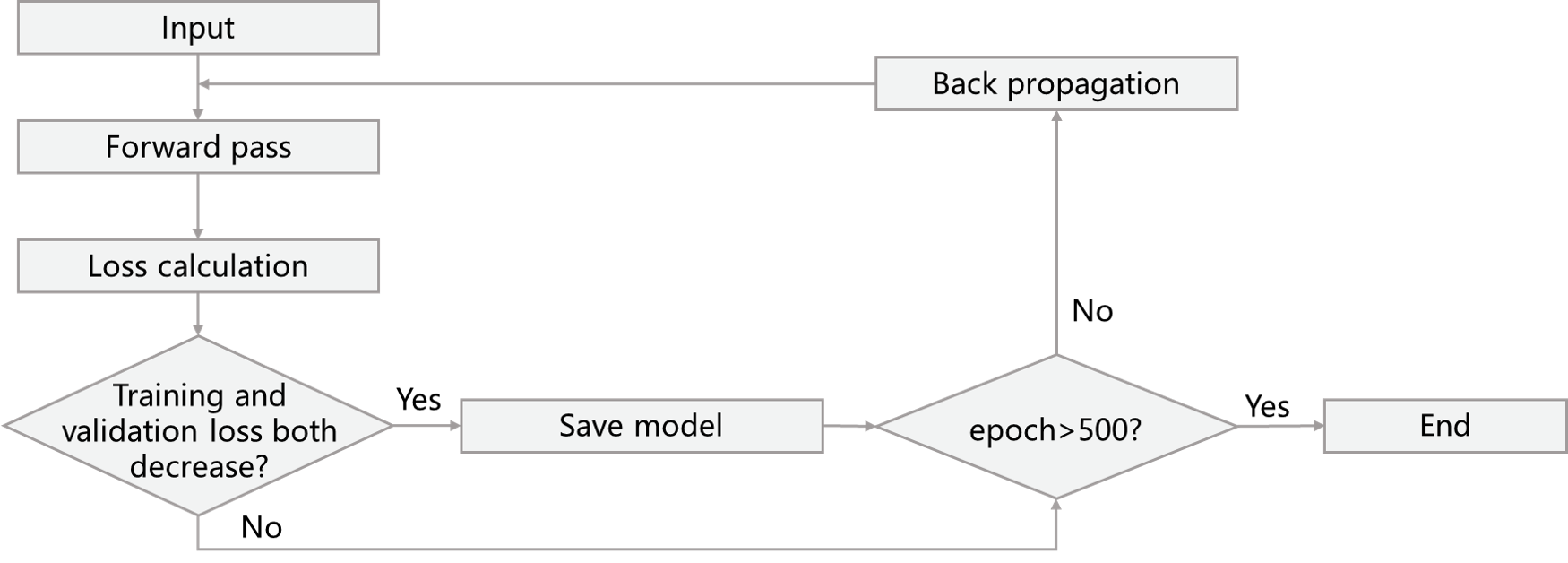}
	\caption{Training strategy of the neural network. The model is saved only when both training and validation losses decrease, and training stops after 500 epochs with the final model corresponding to the relatively lowest combined losses.}
	\label{training_strategy}
\end{figure}

Figure \ref{training_strategy} illustrates the overall training procedure of the neural network. During training, the training loss and the interpolation validation loss are evaluated simultaneously. The model weights are saved only when both losses decrease compared to their previous lowest values. Although the weight updates in the back propagation are carried out only considering the training set, the interpolation validation set is involved in the training process by determining the checkpoints to be saved. Therefore, the final saved model corresponds to the set of weights that achieve the lowest relative values of both training and interpolation validation loss, ensuring generalization performance. In addition, the training process is constrained by a maximum number of epochs, and the training is stopped once the epoch exceeds 500.

\section{Predictive performance of the surrogate model}
\subsection{Effects of residual blocks and neuron numbers on training and validation loss}
Table \ref{loss_comparison} compares the training and validation losses of the FCRN and FCN models across different network structures. For Case 1 (fast ramping), the training loss decreases as the network depth ($L$) or the number of neurons per layer ($H$) increases, indicating that larger networks fit the training data more effectively. Conversely, the validation loss exhibits an increasing trend with larger networks, reflecting a transition towards overfitting. Moreover, for a given network configuration, the validation loss is typically one to two orders of magnitude larger than the training loss. This demonstrates that the model cannot achieve the same level of accuracy on the interpolation validation set as on the training set, even though all input parameters of the validation set fall within the training domain. Furthermore, the FCRN consistently outperforms the FCN on both the training and validation sets in Case 1, demonstrating the effectiveness of the residual structure.

For Case 2 (steady operation), the overall convergence and generalization performance are worse compared to Case 1, accompanied by larger loss fluctuations. This degradation is primarily attributed to the incorporation of $J_{\mathrm{c}}(B)$ relationship, while still relying on $J_{\mathrm{c}}/J_{\mathrm{c0}}$ for normalization, in addition to an increased input dimensionality. Although the FCRN and FCN achieve comparable training losses in this case, the FCRN outperforms the FCN in terms of validation loss across various structures.

\begin{table*}[htbp]
	\caption{Comparison of Training and Validation Loss ($\times 10^{-6}$) Between FCRN and FCN}
	\centering
	\begin{tabular}{l c c c c c c c c}
		\hline
		& \multicolumn{4}{c}{Case 1 (Fast Ramping)} & \multicolumn{4}{c}{Case 2 (Steady Operation)} \\
		\hline
		& \multicolumn{2}{c}{Training} & \multicolumn{2}{c}{Validation} & \multicolumn{2}{c}{Training} & \multicolumn{2}{c}{Validation} \\
		\hline
		Structure & FCRN & FCN & FCRN & FCN & FCRN & FCN & FCRN & FCN \\
		\hline
		6-128  & $1.53$ & $1.47$ & $3.44$ & $6.34$ & $43.3$ & $44.8$ & $136$ & $131$ \\
		6-256  & $0.445$& $0.388$& $4.69$ & $5.86$ & $9.94$ & $29.5$ & $120$ & $165$ \\
		6-512  & $0.230$& $0.360$& $4.67$ & $5.42$ & $12.4$ & $6.65$ & $120$ & $125$ \\
		\hline
		12-128 & $0.561$& $0.603$& $2.19$ & $4.22$ & $15.6$ & $50.2$ & $74.1$& $237$ \\
		12-256 & $0.233$& $0.269$& $2.74$ & $3.93$ & $2.86$ & $1.94$ & $77.3$& $101$ \\
		12-512 & $0.209$& $0.251$& $3.25$ & $3.45$ & $43.2$ & $59.3$ & $84.7$& $140$ \\
		\hline
		24-128 & $0.401$& $20500$& $1.17$ & $20700$& $9.35$ & $19700$& $68.0$& $20200$\\
		24-256 & $0.233$& $20500$& $1.36$ & $20700$& $8.39$ & $19700$& $62.2$& $20200$\\
		24-512 & $0.169$& $20500$& $3.66$ & $20700$& $2.61$ & $19700$& $65.4$& $20200$\\
		\hline
	\end{tabular}
	\label{loss_comparison}
\end{table*}

\begin{figure}[htbp]
	\centering
	\includegraphics[width=0.95\textwidth]{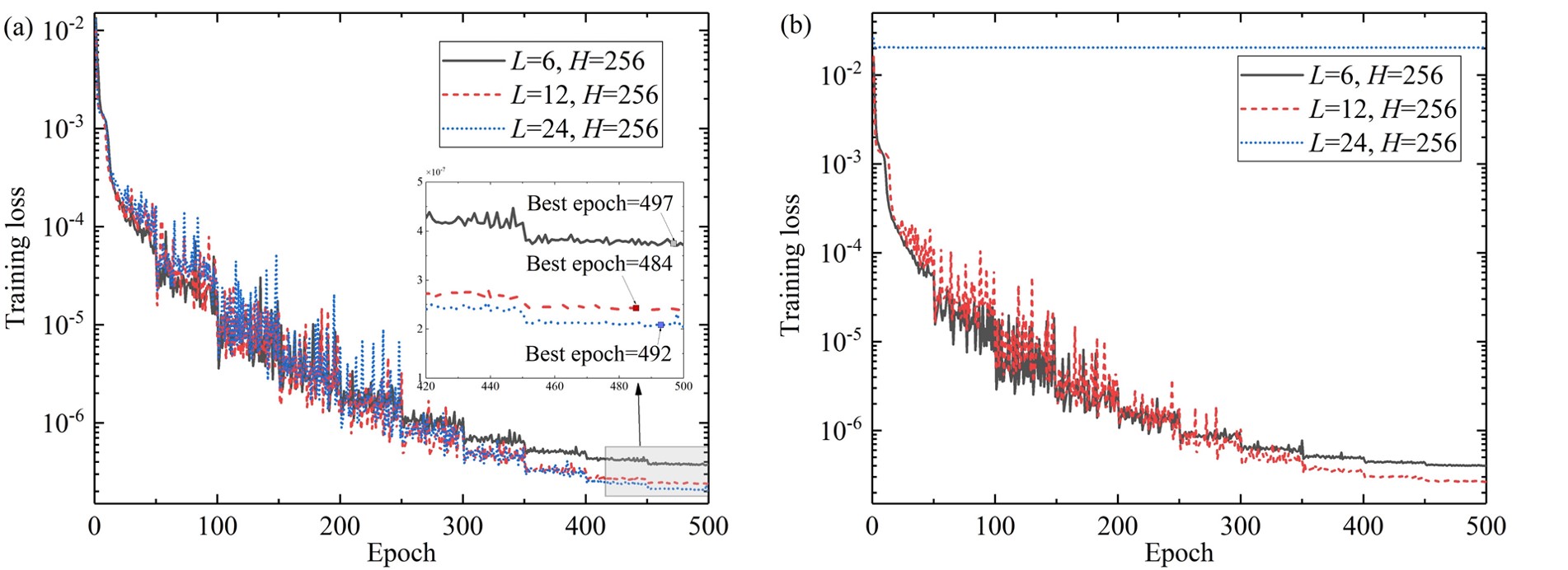}
	\caption{Training loss curves of different network structures for Case 1: (a) residual networks with the best epochs under the adopted training strategy; (b) fully connected networks, where deep networks ($L=24$) fail to converge.}
	\label{loss_compare}
\end{figure}

Fig. \ref{loss_compare} presents the training loss curves for different network structures in Case 1. As shown in Fig. \ref{loss_compare}(a), the FCRN models with $H=256$ successfully converge across varying depths. The inset highlights that the best models are saved near the final epoch, validating the adopted training strategy. Figure \ref{loss_compare}(b) illustrates the corresponding FCN results. While both architectures converge at shallower depths of $L=6$ and $L=12$, the deep FCN with $L=24$ fails to converge, with its training loss maintaining at a high level. This failure demonstrates that without residual connections, deep networks suffer from the vanishing gradient problem during back-propagation. The FCRN effectively mitigates this issue, ensuring stable convergence for deeper architectures.

In conclusion, the FCRN architecture provides better performance for the dataset constructed in this study. Considering the comprehensive performance across both operating scenarios, the FCRN configuration with 24 layers and 256 neurons (24-256) achieves the best balance between training and validation accuracy. Therefore, it is selected as the optimal network architecture for the current density predictions of the HTS solenoid.

\subsection{Extrapolation performances of the FCRN models for Case 1}
\begin{figure}[htbp]
	\centering
	\includegraphics[width=0.65 \textwidth]{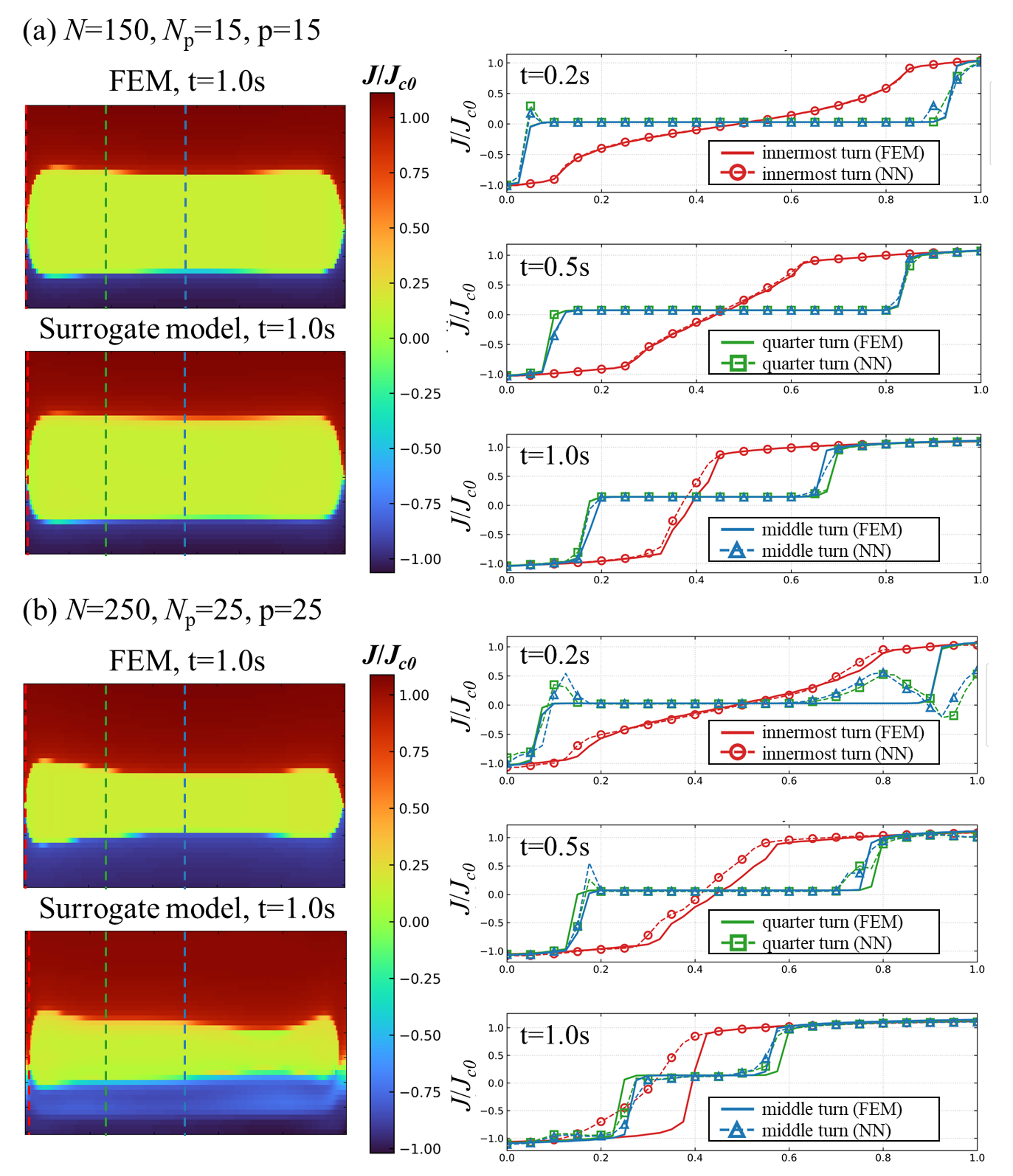}
	\caption{Comparison of current density distributions between the FEM and the surrogate model for Case 1: (a) $N=150$, $N_{\mathrm{p}}=15$, $p=15$ and (b) $N=250$, $N_{\mathrm{p}}=25$, $p=25$. Left contours display the current density at the final ramping moment. Right figures compare the current density profiles along the tape width for the innermost, quarter, and middle turns at different time steps. Solid lines and dashed lines with markers denote the FEM solutions and the NN predictions, respectively.}
	\label{Jz_compare_case1}
\end{figure}

After training, the extrapolation capability of the surrogate models is of particular concern, as it is important for constructing predictive models of large-scale superconducting magnets from relatively small datasets. To verify this capability, the current density distributions predicted by the optimal FCRN structure ($L=24$ and $H=256$) are compared with the FEM results under different extrapolation ratios in Case 1. Fig. \ref{Jz_compare_case1} (a) and (b) presents the comparison for the uppermost pancake coil with both the 2D cross-sectional contours and the 1D normalized current density profiles for the innermost, quarter, and middle turns. As shown in Fig. \ref{Jz_compare_case1}(a), for the configuration with $N=150$ and $N_{\mathrm{p}}=15$ (representing a 50\% extrapolation ratio relative to the training set boundaries), the NN predictions closely match the FEM results. The 1D profiles also reflect this high consistency at various ramping stages ($t=0.2\mathrm{s}$, $0.5\mathrm{s}$, and $1.0\mathrm{s}$). This indicates that within a relatively small extrapolation range, the surrogate model can still achieve reasonably accurate predictions, even without explicit physics-constrained mechanisms during extrapolation.

However, as the extrapolation ratio increases to 150\% ($N=250$, $N_{\mathrm{p}}=25$), large discrepancies occur. As shown in Fig. \ref{Jz_compare_case1}(b), the NN prediction deviates from the FEM result, particularly at the lower area of the pancake. The 1D normalized current density profiles clearly illustrate these significant deviations, demonstrating that the model fails to accurately predict the distributions at extreme extrapolation scales.

To investigate the extrapolation limits of the surrogate model, the hysteresis loss power during the fast ramping process is calculated and compared. The magnetization loss is given by (\ref{eq:loss}):
\begin{equation}
	P=\int_{t}^{} E_{\varphi} \cdot J_{\varphi} dt.
	\label{eq:loss}
\end{equation}

\begin{figure}[htbp]
	\centering
	\includegraphics[width=1.0\textwidth]{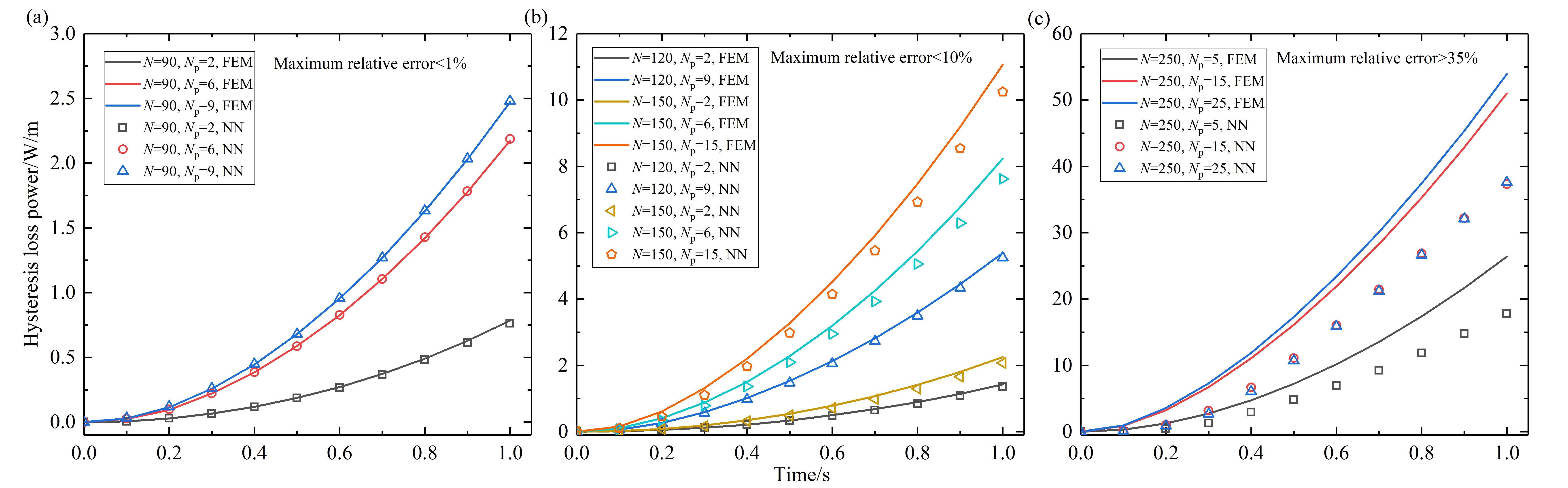}
	\caption{Comparison of the hysteresis loss power for Case 1 between FEM results and surrogate model predictions: (a) interpolation within the training set, (b) extrapolation with a ratio of 50\%, and (c) extrapolation with a ratio of 150\%. Solid lines represent the FEM solutions, while symbols denote the NN predictions.}
	\label{loss_dif}
\end{figure}

The hysteresis loss power curves obtained from the FEM and the surrogate model are compared under different extrapolation scenarios. Fig. \ref{loss_dif}(a) displays the interpolation results within the training set ($N=90$, $N_{\mathrm{p}} \le 9$), where the surrogate model achieves high accuracy with a maximum relative error of less than 1\%. For the limited extrapolation cases (extrapolation ratio up to 50\%, e.g., $N=120, 150$ and $N_{\mathrm{p}} \le 15$), as shown in Fig. \ref{loss_dif}(b), the predicted loss power curves still align well with the FEM results, keeping the maximum relative error below 10\%. This indicates that the model provides reliable predictions for extrapolation within $N_{\mathrm{p}} \le 15$ and $N \le 150$. Conversely, when the geometric parameters are extrapolated to much larger ranges (extrapolation ratio of 150\%, $N=250$ and $N_{\mathrm{p}} \le 25$), as illustrated in Fig. \ref{loss_dif}(c), the relative error of the hysteresis loss power exceeds 35\%, indicating a significant error of predictive results. In conclusion, with an appropriate network architecture and hyperparameter configuration, a training set with relatively small parameter ranges can effectively provide accurate predictions for conditions moderately beyond the training domain.

Table \ref{time_compare} compares the inference time of the surrogate model with the computation time of FEM models for Case 1. The results clearly show that the surrogate model achieves predictions several orders of magnitude faster than FEM. Even when accounting for the training time, which is approximately 5000$\,$s, the total time still remains lower than that of FEM simulations, which demonstrates that the proposed surrogate model not only provides accurate extrapolation performance but also offers a substantial advantage in computational efficiency.

\begin{table}[htbp]
	\caption{Comparison of model inference time and FEM computation time}
	\centering
	\begin{tabular}{l c c c}
		\hline
		Input parameter & FEM computation time & Model inference time \\
		\hline
		$N=100$, $N_{\mathrm{p}}=10$ & 1h13min & 0.107s\\ 
		$N=150$, $N_{\mathrm{p}}=15$ & 2h44min & 0.176s\\  
		$N=250$, $N_{\mathrm{p}}=25$ & 11h6min & 0.366s\\
		\hline
	\end{tabular}
	\label{time_compare}
\end{table}

\subsection{Extrapolation performances of the FCRN models for Case 2}
To evaluate the extrapolation capability of the surrogate model for Case 2 (steady operation), the central magnetic field is chosen as an validation index. The surrogate model first predicts the spatial distribution of the normalized current density $J_{\varphi}/J_{\mathrm{c0}}$. Then the central magnetic field $B_{\mathrm{0}}$ is calculated by the Biot-Savart law.

\begin{figure}[htbp]
	\centering
	\includegraphics[width=0.6\textwidth]{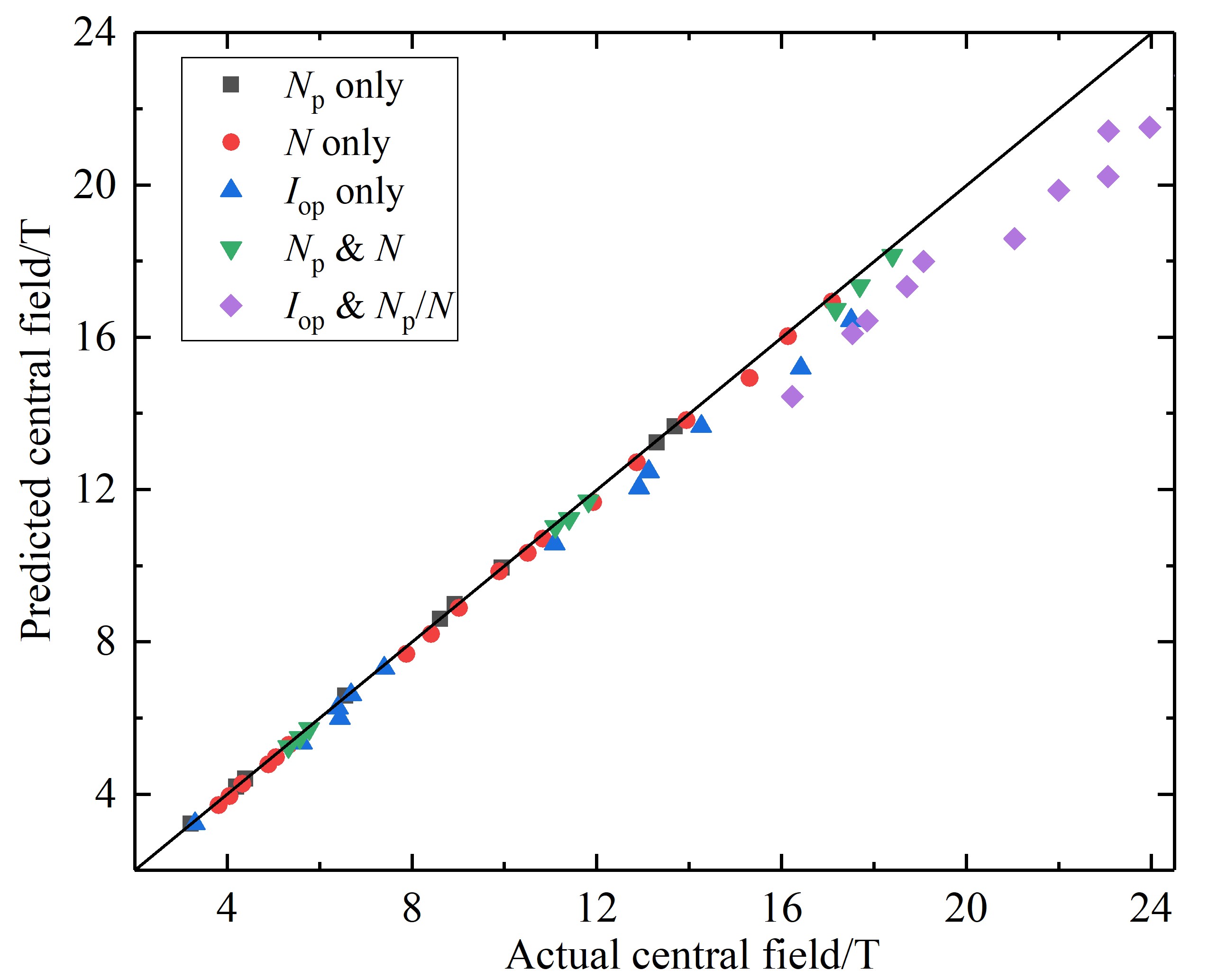}
	\caption{Comparison of the central magnetic field calculated by the surrogate model predictions against the FEM results for Case 2 (steady operation). The scatter plot evaluates the model's extrapolation performance across various scenarios. The solid black line indicates the ideal agreement ($y=x$).}
	\label{field_compare_case2}
\end{figure}

Fig. \ref{field_compare_case2} compares the central magnetic fields derived from the surrogate model and the FEM calculations. The evaluation includes following extrapolation scenarios: extrapolating the number of pancakes to $N_{\mathrm{p}}=10$ (with training set maximum of 8), extrapolating the number of turns to $N=400$ (with training set maximum of 300), extrapolating the operating current to $I_{\mathrm{op}}=300\,\mathrm{A}$ (with training set maximum of $225\,\mathrm{A}$), and combinations of these parameters. As observed, the predicted central magnetic fields align well with the FEM results when only geometric parameters ($N$ or $N_{\mathrm{p}}$) are extrapolated, showing a low average relative error of 1.2\%. This indicates that the surrogate model maintains robust extrapolation performance for geometric variations, which is consistent with the conclusions drawn in Case 1. 

However, when the operating current $I_{\mathrm{op}}$ is extrapolated, obvious discrepancies arise. The average error increases to 4.4\% for independent $I_{\mathrm{op}}$ extrapolation and further to 9.0\% for combined extrapolations involving $I_{\mathrm{op}}$ and geometric parameters. Consequently, the surrogate model exhibits limited generalization capacity when extrapolating the operating current beyond the training domain.

\begin{figure}[htbp]
	\centering
	\includegraphics[width=0.65\textwidth]{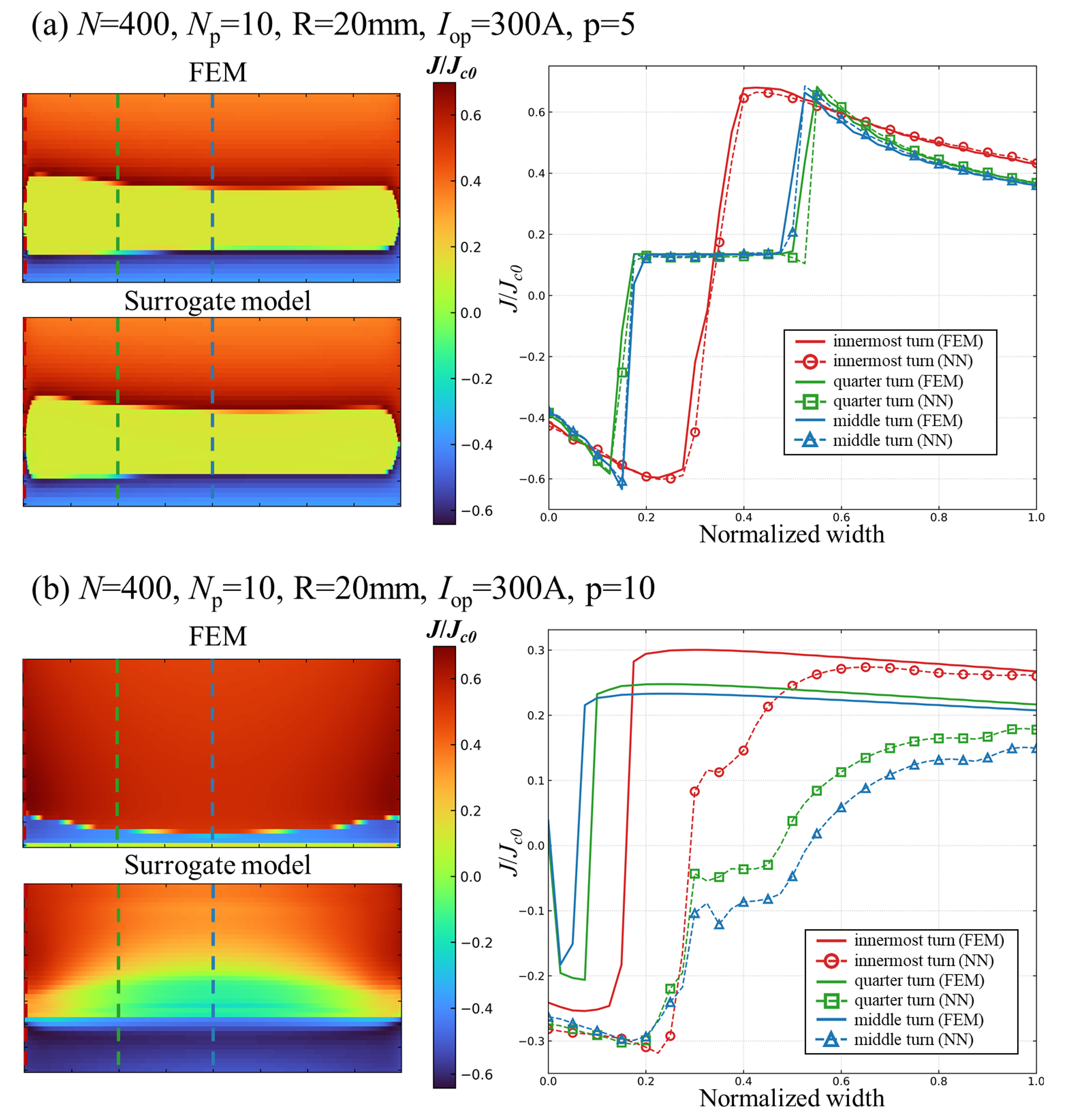}
	\caption{Comparison of current density distributions between the FEM and the surrogate model for Case 2 (steady-state) when $N=400$, $N_{\mathrm{p}}=10$, $R=20\,\mathrm{mm}$, and $I_{\mathrm{op}}=300\,\mathrm{A}$: (a) the middle pancake ($p=5$) and (b) the uppermost pancake ($p=10$). The left contours display the 2D cross-sectional current density and the right figures compare the 1D normalized current density profiles along the tape width for the innermost, quarter, and middle turns. Solid lines and dashed lines with open markers denote the FEM solutions and the NN predictions, respectively.}
	\label{Jz_compare_case2}
\end{figure}

To further illustrate the reasons for the extrapolation errors at high operating currents, Fig. \ref{Jz_compare_case2} show the current density distributions for the middle pancake ($p=5$) and the uppermost pancake ($p=10$) in a combined extrapolation scenario ($N=400$, $N_{\mathrm{p}}=10$, $R=20\,\mathrm{mm}$, and $I_{\mathrm{op}}=300\,\mathrm{A}$). As shown in Fig. \ref{Jz_compare_case2}(a), the surrogate model still accurately predicts the current density distribution in the middle pancake. However, significant deviations occur in the uppermost pancake, as depicted in Fig. \ref{Jz_compare_case2}(b). 
The tapes in the uppermost pancake experience a large perpendicular magnetic field, which severely suppresses the local critical current density $J_{\mathrm{c}}(B)$, resulting in a fully penetration stage. The training dataset for Case 2, however, does not contain samples that capture how the current density distribution evolves when the operating current is further increased under such conditions. 
Therefore, it fails to extrapolate the underlying nonlinear penetration dynamics. For the middle and lower pancakes, the perpendicular magnetic field is relatively small, leaving the tapes only partially penetrated. Thus, the network can still rely on the penetration laws learned from the training domain to provide reliable predictions.

\subsection{Rapid magnet design based on surrogate model for Case 2}
Once the surrogate model is fully trained and its generalization boundaries are validated, its most significant advantage lies in the ability to perform rapid parameter sweeps for practical magnet design. Traditionally, optimizing the design of a high-temperature superconducting (HTS) solenoid with FEM models is time-consuming due to the need for repetitive simulations. In contrast, the surrogate model enables rapid predictions. Furthermore, its extrapolation capability eliminates the need for a large training dataset, and a model trained on a smaller parameter range can reliably optimize designs within a broader domain.

\begin{figure}[htbp]
	\centering
	\includegraphics[width=0.4\textwidth]{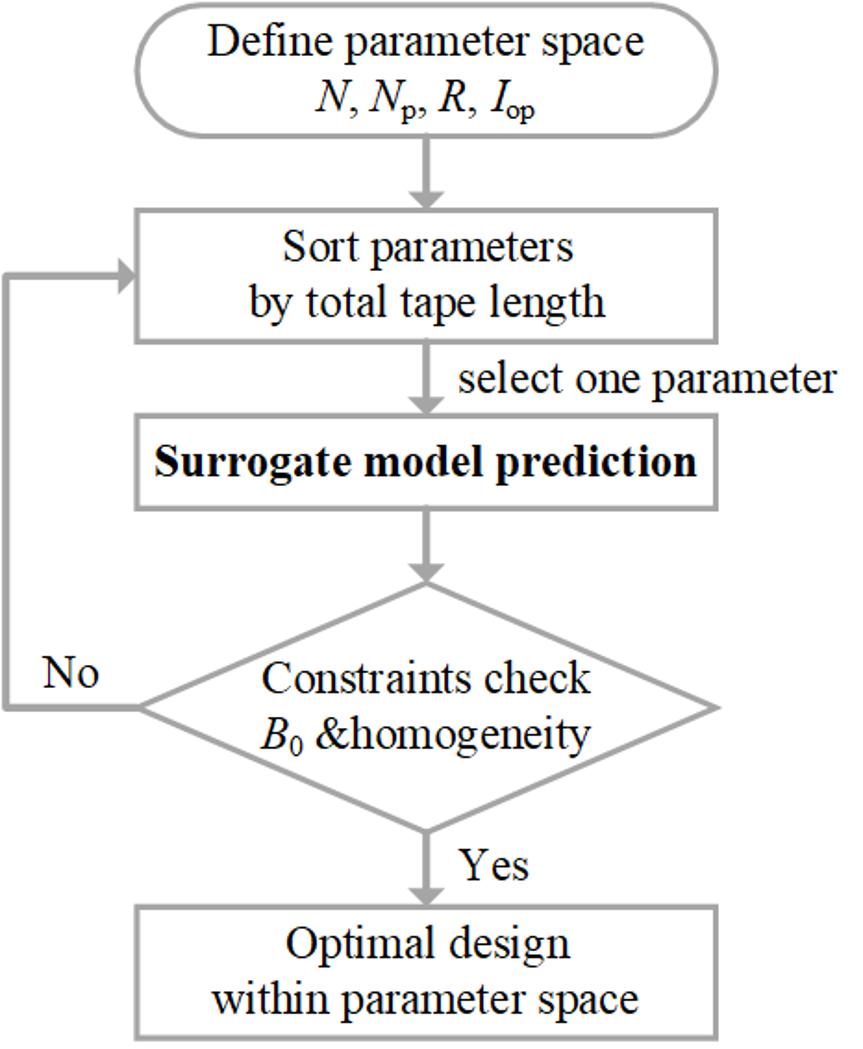}
	\caption{Flowchart of the optimization procedure for the HTS solenoid design based on the surrogate model trained for Case 2 (steady operation).}
	\label{Magnet_design}
\end{figure}

To demonstrate this, the surrogate model trained for Case 2 (steady operation) is deployed to design an optimal HTS solenoid. The design objective is to minimize the total tape consumption while satisfying two electromagnetic constraints: the central magnetic field ($B_{\mathrm{0}}$) exceeds 16\:T, and the field homogeneity within 20\:mm range along the central axis is less than 1\%. Based on the previous discussion, the parameter search space is defined as $N \le 400$, $N_{\mathrm{p}} \le 10$, $R \in [10, 20]$ mm, and $I_{\mathrm{op}} \le 225$ A. Fig. \ref{Magnet_design} illustrates the flowchart of the proposed optimization procedure. Because the forward prediction of the neural network is extremely fast, a direct heuristic search strategy is adopted. First, all possible discrete parameter combinations within the defined search space are generated. These candidate designs are then sorted in ascending order based on their required total tape length. Subsequently, the surrogate model sequentially predicts the spatial current density distributions for these sorted candidates. The predicted current density is then integrated via the Biot-Savart law to check whether the $B_0$ and homogeneity constraints are met. Because the candidates are evaluated from the shortest tape length to the longest, the very first parameter combination that satisfies both electromagnetic constraints is guaranteed to be the optimal design. Sweeping the entire parameter space and finalizing this optimization procedure took about 3 minutes on the training work station.

\begin{figure}[htbp]
	\centering
	\includegraphics[width=0.65\textwidth]{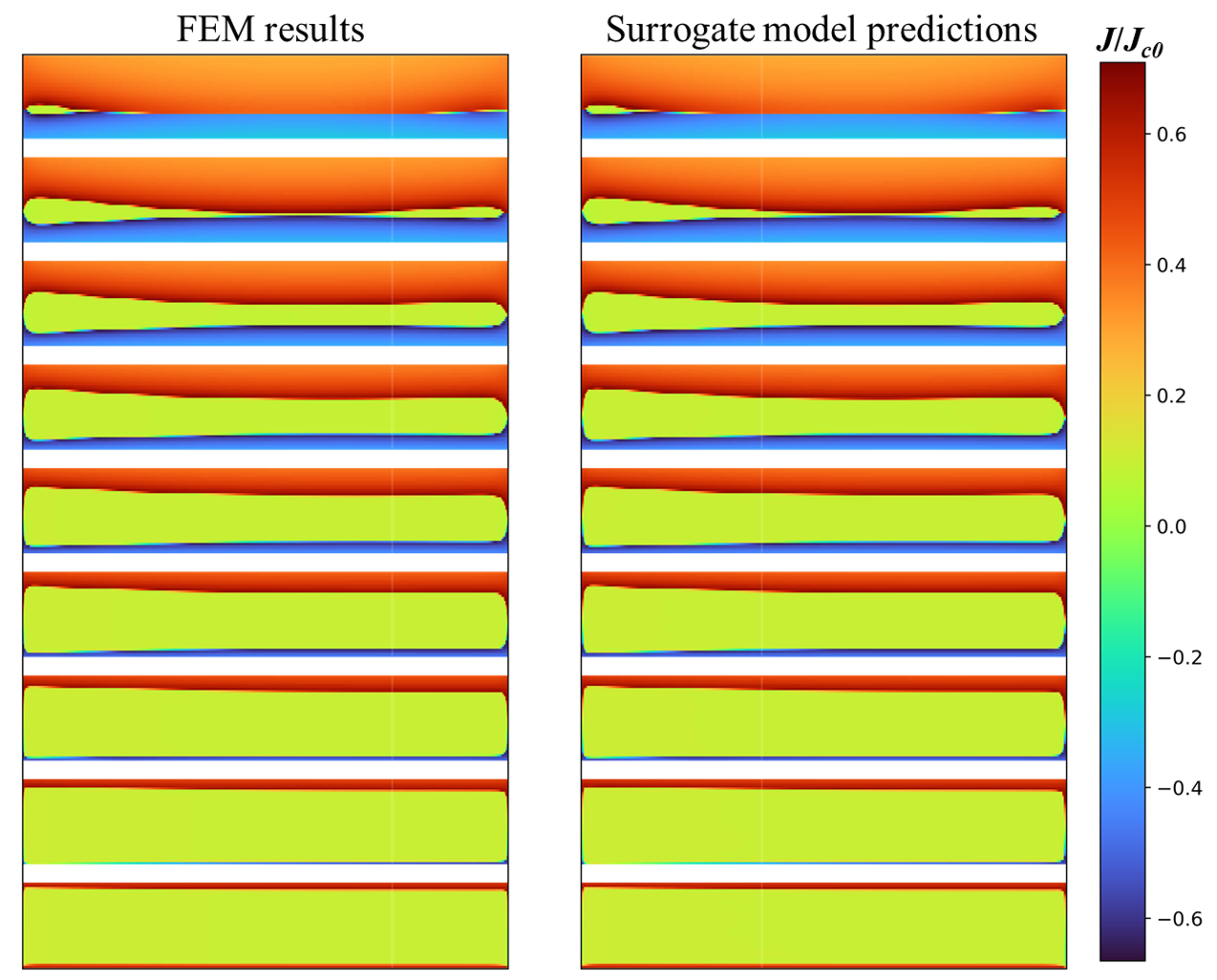}
	\caption{Comparison of the normalized current density distributions between the FEM results and the surrogate model predictions for the optimal HTS solenoid design given by the optimization procedure, with the parameter combination of $N=360$, $N_{\mathrm{p}}=9$, $R=10$ mm, and $I_{\mathrm{op}}=222$ A.}
	\label{Magnet_design_case}
\end{figure}

The optimization procedure yielded an optimal HTS solenoid design with the parameter combination of $N=360$, $N_{\mathrm{p}}=9$, $R=10$ mm, and $I_{\mathrm{op}}=222$ A. This optimal design falls into the extrapolation region for the geometric parameters. To  verify this result, a full FEM simulation was conducted with this optimal parameters. Fig. \ref{Magnet_design_case} presents the comparison of the normalized current density distributions between the FEM results and the surrogate model predictions. Despite the extrapolation, the neural network's predictions closely match the FEM results across all pancakes. The calculated central magnetic field from the surrogate model's prediction (16.02 T) is in good agreement with FEM results (16.04 T). These results validate the feasibility, accuracy, and efficiency of employing the surrogate model for rapid design and optimization of large-scale HTS magnets.

\section{Conclusion}
In this work, a surrogate model based on a fully connected residual network (FCRN) was developed to predict the current density distribution in high-temperature superconducting solenoids. By constructing datasets through FEM simulations with the $T$–$A$ formulation, the model was trained under fast ramping (Case 1) and steady-state (Case 2) conditions, and its extrapolation performance was examined in detail.

The results demonstrated that the FCRN architecture outperforms FCNs. In Case 1, the FCRN achieved lower training and validation losses, and could reliably predict ramping hysteresis loss with parameters extrapolation up to 50\% while keeping the relative error below 10\%. For Case 2, incorporating $J_{\mathrm{c}}(B)$ dependence and additional input dimensions degraded the overall fitting performance, but the FCRN still had lower validation losses compared with FCN. Due to the dataset lacked current evolution dynamics under full tape penetration, the model showed limited current extrapolation capability in Case 2. However, it maintained high accuracy for geometric extrapolations, achieving an average central magnetic field prediction error of 1.2\%. 

The surrogate model trained for Case 2 was deployed for rapid magnet design. By constraining the central magnetic field and axial field homogeneity, the model identified the optimal design within the parameter search space. This confirms that surrogate models offer a promising and computationally efficient tool for the intelligent design of high-field HTS magnets. Future work will focus on constructing more representative datasets and incorporating physical constraint losses to further enhance the generalization and extrapolation performances of the model.

\ack{This work was supported in part by the National MCF Energy R\&D Program under Grant No. 2022YFE03150103, and in part by the National Natural Science Foundation of China (NSFC) under Grant No.52277026}

\end{document}